# MultiEarth 2022 - The Champion Solution for Image-to-Image Translation Challenge via Generation Models

Yuchuan Gou
PAII Inc.
Palo Alto, CA, USA
plain1994@gmail.com

Bo Peng
PAII Inc.
Palo Alto, CA, USA
bpeng.paii@gmail.com

Hongchen Liu
Pingan Tech Inc.
Shenzhen, Guangdong, China
liuhongchen453@pingan.com.cn

Hang Zhou
PAII Inc.
Palo Alto, CA, USA
joeyzhou1984@gmail.com

Jui-Hsin Lai
PAII Inc.
Palo Alto, CA, USA
juihsin.lai@gmail.com

## Abstract

*The MultiEarth 2022 Image-to-Image Translation challenge provides a well-constrained test bed for generating the corresponding RGB Sentinel-2 imagery with the given Sentinel-1 VV & VH imagery. In this challenge, we designed various generation models and found the SPADE [1] and pix2pixHD [2] models could perform our best results. In our self-evaluation, the SPADE-2 model with L1-loss can achieve 0.02194 MAE score and 31.092 PSNR dB. In our final submission, the best model can achieve 0.02795 MAE score ranked No.1 on the leader board.*

## 1. Introduction

Recent years have seen rapid developments of earth observation satellites, with more and more optical and Synthetic Aperture Radar (SAR) data become available. While optical data are often obscured by cloud, haze, smoke, shadow (mountain or cloud), low light condition, direct reflected light, and other issues, SAR has the benefit of being insensitive to weather or lighting situations, due to its nature of being an active microwave radar. But this also causes one of the major drawbacks of SAR, the data is hard to be visualized, explained, and understood by human operators without appropriate training and experiences with SAR data.

Image-to-Image Translation could be a possible approach for solving this problem, by transferring SAR images in the radar reflected signal intensity domain to the more commonly known as optical spectral domain, or even the simpler RGB color domain. The MultiEarth 2022 Image-to-Image Translation challenge provides a well-constrained test bed, with very meaningful downstream applications, for developing such a method. After initial exploratory data analysis, we notice that: 1) a lot of training data provided are hindered with a large percentage of cloud and would significantly decrease model performance if used; 2) generative style model could work better than regression style model for this task due to the substantial modality difference; 3) a combination of different models trained with different subset of data would be able to provide a better coverage of the targeted RGB domain based on the given SAR data. Based on these findings, we design our modeling Methodology and conduct a series of Experiments, which are presented in the following sections.

## 2. Methodology

This work builds on the semantic image synthesis model, SPADE [1], for the translation from synthetic-aperture radar (SAR) imagery to electro-optical (EO) Sentinel-2 imagery. The SPADE uses spatially-adaptive transformation to modulate the activation in normalization layers by virtue of the input semantic layout. Such a simple but effective layer proves to be effective for synthesizing photo-realistic images if the image semantic layout is available. Instead of using the semantic layout as the input, this paper feeds into the model with Sentinel-1 VV & VH imagery and generates the corresponding RGB Sentinel-2 imagery.

To leverage the power of different models, this paper tries the other effective image-to-image translation model, pix2pixHD [2], for generating Sentinel-2 imagery given the Sentinel-1 imagery. The pix2pixHD model uses a new multi-scale generator and discriminator as well as a novel adversarial loss for synthesizing visually-appealing high-resolution images of size 2048x1024 given the input image semantic layout.

Both the SPADE and pix2pixHD behave as a variant of the conditional generative adversarial net (cGAN) but generate images with much high spatial resolution. Therefore, this paper leverages both models for translating Sentinel-1 images of size 256x256 to Sentinel-2 images of the same size.

We have tried GAN Loss, L1 loss, and different weighted loss combinations on these Image Translation models. We found that, with L1 loss for training, we can get better MAE / PSNR performance on our evaluation.

## 3. Experiments

### 3.1. Datasets

During our initial exploratory data analysis of the provided training pairs of S1-SAR vs S2-RGB images, we realized there are a large amount of S2-RGB images with large percentage of nodata pixels, or cloudy pixels. To avoid including those images with lower quality as target images during training, we populated all given S2-RGB images, computed the nodata ratio and cloudy ratio based on the given companion QA60 band for each image. Then we filtered the training dataset by only keeping S2-RGB images with both nodata ratio and cloudy ratio at 0. The result is our Dataset-1, containing 81,391 paired S1-SAR/S2-RGB images.

We randomly picked up image pairs from Dataset-1 and notice there are still lots of images with remaining cloudy pixels. It is well-known in the remote sensing community that QA60 band's cloud mask capability is of relatively lower quality, especially compared with its counterpart in the SCL band. Due to the fact that we were not provided the source id for the images, it would be too complex and quite time-consuming to retrieve the corresponding SCL band for those training images. Thus we developed a simple heuristic based method for per-pixel cloud detection, based on difference between brightness and saturation, following [3]. We further filtered Dataset-1 based on the cloud ratio derived from the heuristic-based cloud mask, and kept 35,587 cloud-free pairs of S1-SAR/S2-RGB images with cloud ratio of 0 as our Dataset-2.

For self-evaluation during training, we set aside 1,000 pairs from Dataset-2. Since Dataset-2 is almost cloudy-free, we found that evaluation on data samples from this dataset much more accurate for MAE and PSNR metrics.

### 3.2. Results and Submissions

We trained models (SPADE and pix2pixHD) with different loss and dataset, with the result reported in Table 1. The self-evaluation shows the SPADE model with L1-loss can achieve the best results in terms of MAE and PSNR metrics. We also noticed that visually, SPADE models trained with GAN-loss tend to provide more variances and spatial details in the resulting RGB images. Thus we include different variants of models in our submissions, in order to provide more comprehensive coverage of the possible solution space. We performed two submissions with different combinations of model outputs, with the result shown in Table 2.

| Model | Loss | Dataset | MAE | PSNR |
|---|---|---|---|---|
| SPADE-1 | GAN + 1000 * L1 | Dataset-1 | 0.02447 | 30.1179 |
| SPADE-2 | L1 | Dataset-1 | **0.02194** | **31.0926** |
| SPADE-3 | 100 * L1 | Dataset-1 | 0.02212 | 31.0589 |
| pix2pixHD-1 | L1 | Dataset-1 | 0.02258 | 30.9797 |
| SPADE-4 | GAN + 1000 * L1 | Dataset-2 | 0.02384 | 30.0793 |

Table 1. Model training experiments with different loss settings and dataset.

| Submission | Models | MAE |
|---|---|---|
| Submission1 | SPADE-1, SPADE-2, SPADE-3 | 0.0344979 |
| Submission2 | SPADE-4, SPADE-2, pix2pixHD-1 | **0.0279459** |

Table 2. Submission Results

## 4. Conclusion

The generation model could work better than regression style model for the image-to-image translation tasks due to the substantial modality difference. Our experiments show the generation models with GAN-loss can produce more visually intensive and sharpened images than the models with only L1-loss. However, the L1-loss models can achieve higher MAE and PSNR scores that we preferred the models with L1-loss for the final submission. Also, we found lots of training data provided are hindered by a large percentage of cloud and would significantly decrease model performance. We trained the models on the processing dataset with removing cloud pixels with our own cloud detection algorithm.